# Title

Harnessing Plastic Deformation for Continuous Passive Stability in Soft Grippers

# Authors

Junlong Xiao [a], Yaoqiang Pan [a], Xuan Zhang [a], Michael Yu Wang [b], Chao Chen [a]*

# Affiliations

[a] Faculty of Engineering, Monash University, Clayton, VIC 3800, Australia

[b] Michael Yu Wang. School of Engineering, Great Bay University, Songshan Lake, Dongguan, Guangdong, China

* Address correspondence to: chao.chen@monash.edu

# Abstract

Soft robots offer compliance and adaptability for interacting with complex environments, but their low structural stiffness can compromise stability under dynamic disturbances. Here, we present a continuously stable structure through plastic deformation (CSSPD), a hybrid metal–elastomer architecture that enables passive shape retention without continuous energy input. By deliberately exploiting plastic deformation of a kirigami-patterned metal layer, CSSPD establishes new equilibrium configurations while increasing structural stiffness and resistance to dynamic disturbances. A theoretical model is developed to characterize the deformation and stability of the hybrid structure and to guide the geometric design of the metal layer. We further introduce a bioinspired paw pad that provides top-side support for enhanced grasp stability and functions as an integrated sensing interface for rapid object acquisition. A Telescopic Soft Pressing Mechanism mechanically (TSPM) couples paw-pad motion with tendon actuation. Experiments demonstrate passive holding forces of up to 16 N without continuous energy input, comparable to pneumatic actuation at 0.3 MPa, and stable grasping under pulsed accelerations of up to 400 m/s $^2$when combined with pneumatic actuation. Energy-free perching on tree branches further demonstrates the potential of the proposed composite architecture for stable and energy-efficient manipulation. Project videos and additional details are available at : https://github.com/JunlongXiao/CSSPD_supplementary_materials

# 1. Introduction

Soft robots, particularly soft grippers, have attracted increasing interest for industrial and field applications because their intrinsic compliance enables safe and adaptable physical interactions [1, 2]. However, the same compliance that facilitates interaction can compromise grasp stability during high-speed manipulation. In applications such as field harvesting and industrial assembly, reducing manipulation time requires rapid acceleration and deceleration of the robotic arm and gripper, which generates inertial forces that can cause grasped objects to shift, swing, or detach. At the same time, energy-efficient operation requires energy to be concentrated on grasp formation and release rather than continuously expended during motion. These competing requirements motivate the development of soft grippers that can maintain stable configurations and grasping forces without continuous actuation.

Soft materials and structures provide the flexibility required for adaptive manipulation, but often at the expense of structural stability. To address this trade-off, both active and passive stabilization strategies have been developed. Active approaches include variable-stiffness mechanisms based on jamming [3-5], shape memory polymers (SMPs) [6-9], thermally actuated materials [10-12], magnetic fields [13], electric fields [14], and combined optical-electric stimulation [15]. These approaches can dynamically regulate stiffness or configuration, but maintaining the reconfigured state generally requires continued energy input or active control. They also introduce additional actuators, control complexity, and, in some cases, response delays, which can limit their suitability for energy-constrained and rapidly moving robotic systems.

In contrast to active approaches, passive stabilization relies on structures that require energy only to initiate a change in state, without the need for continuous input to maintain it, which significantly reduces energy consumption and extends the operational life of the robot [16]. This approach offers a distinct and reliable solution for environments with limited or unreliable energy sources, such as those powered by solar panels, batteries, or operating in conditions where consistent electrical power cannot be ensured [17, 18].

Passive stable structures based on geometric design provide an alternative route to energy-free state retention. Bistable structures [19-28], for example, possess two local minima in their potential-energy landscapes and can transition between stable configurations after an external input overcomes the energy barrier. By combining multiple bistable units, multistable structures with multiple equilibrium states can be constructed [29-34]. From a mechanical perspective, a stable grasp can likewise be associated with a local minimum of the system's potential energy [35, 36]. However, the equilibrium configurations of bistable and multistable systems remain discrete and finite. Consequently, achieving continuously adjustable stable configurations requires either an increasing number of discrete states or an alternative stabilization mechanism [37, 38]. Ratchet-inspired self-locking mechanisms [39-42] and snap-fit joints [43, 44] provide additional passive locking strategies, but their stable configurations are likewise determined by discrete geometric features, such as teeth or snap positions, which can limit continuous adjustability.

Passive continuous stability refers to the ability of a structure to retain a continuously adjustable range of equilibrium configurations after unloading, rather than only a finite set of discrete states. Existing approaches have achieved this behavior through structural friction or material-level memory and plasticity. Friction-based mechanisms, such as multilayer rotational jamming [45-47], provide passive locking through frictional interfaces but typically require multiple structural components and can increase system size and complexity. In our previous work, we exploited self-locking friction in a gooseneck structure to achieve spatially stable configurations [48, 49]. However, its helical geometry restricts the range of configurations and applications. Material-based approaches, including shape-memory alloys (SMAs) [50, 51], can retain deformed configurations but are associated with slow response, hysteresis, and limited passive stiffness. Continuous stability has also been demonstrated using the plastic deformation of low-temperature metals [52]. In that approach, gravitational or pneumatic loading deforms the metal after heating, while the elastic recovery of the surrounding soft material restores the structure following the metal's phase transition. However, the phase transition requires substantial thermal energy and introduces additional response time. More importantly, the mechanical interplay between plastic deformation and elastic recovery, which is central to achieving continuously stable configurations, remains insufficiently characterized.

In this work, we introduce a soft gripper incorporating a continuously stable structure through plastic deformation (CSSPD), which combines the compliance and adaptability of soft robotics with persistent passive stability (Fig. 1 and Movie S1). The CSSPD employs kirigami-structured metal sheets embedded within the soft fingers to exploit plastic deformation and elastic recovery for continuously adjustable equilibrium configurations. This enables the gripper to retain its bent configuration and grasping force after actuation is removed, while the embedded metal structure provides additional stiffness for resisting inertial disturbances. We further develop a bioinspired paw pad with integrated sensing capability and a Telescopic Soft Pressing Mechanism (TSPM). The TSPM mechanically couples paw-pad extension and retraction with tendon actuation, allowing the paw pad to provide additional top-side support during grasping while enabling rapid object acquisition through sensing. A theoretical model is established to characterize the deformation and stability of the kirigami structure and guide its geometric optimization. Experiments demonstrate passive holding forces of up to 16 N without continuous energy input and stable operation under pulsed accelerations of up to 400 $m/s^2$ when combined with pneumatic actuation. The gripper also demonstrates prolonged passive perching on tree branches without power, highlighting its potential for energy-efficient and dynamically stable manipulation in industrial, field, and mobile robotic applications.

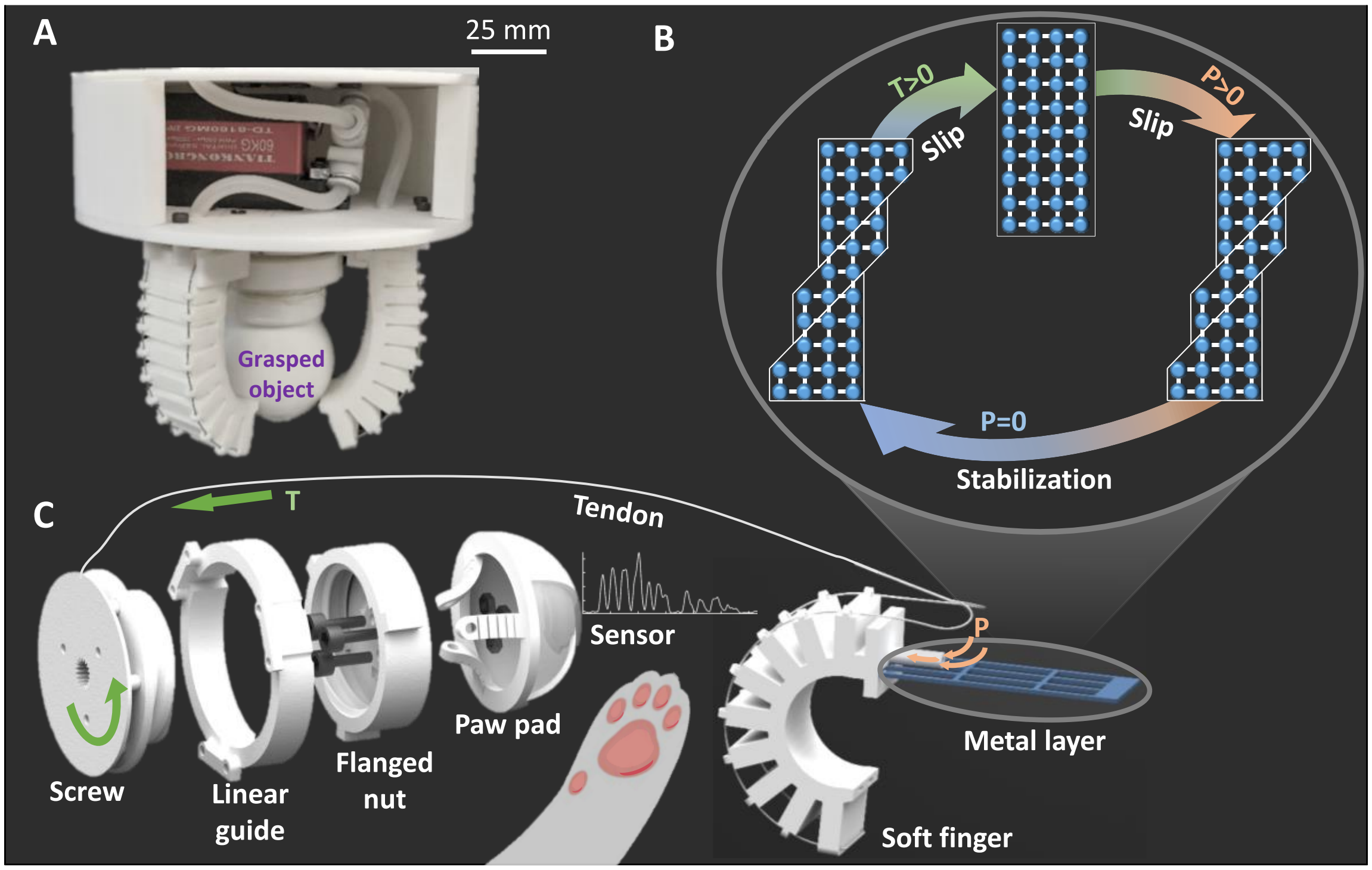

**Fig. 1. Design and working principles of CSSPD mechanism.** (A) A soft gripper designed based on the CSSPD. (B) The soft fingers are actuated by both pneumatic pressure and tendon-driven mechanisms. The CSSPD principle relies on the plastic deformation of the internal metal layer within the soft fingers. (C) Schematic illustration of the components of the Telescopic Soft Pressing Mechanism (TSPM), which enhances the system's grasping stability.

# 2. Design and Modeling

In this section, we present the operating mechanism of the CSSPD gripper and establish a static model. The model is validated through experiments, demonstrating its accuracy and reliability, and is subsequently employed to optimize the design of the metal layer.

## 2.1. Design of the CSSPD Gripper

As shown in Fig. 1A, the gripper consists of a Telescopic Soft Pressing Mechanism (TSPM), a motor, and three continuously stable soft fingers. The CSSPD mechanism is achieved by embedding a kirigami-patterned metal layer inside the soft fingers. When the applied stress exceeds the yield strength, the metal undergoes plastic deformation, primarily through dislocation slip and, under specific conditions, deformation twinning. Dislocation slip is generally the dominant mechanism, whereas twinning becomes more significant when slip is constrained, such as under limited available slip systems, low temperatures, or high strain rates. Importantly, plastic deformation does not provide elastic energy storage for maintaining the deformed configuration. Instead, it produces permanent plastic strain that irreversibly changes the material geometry.

After unloading, the resulting residual deformation allows the structure to retain a new equilibrium configuration, without requiring continuous external energy input to maintain the deformed state.

As shown in Fig. 1B, the soft fingers are actuated by both pneumatic and tendon-driven mechanisms. When the fingers are inflated (P>0), the metal layer bends, causing atomic bonds within the crystal slip planes to break, allowing relative atomic motion before reforming. Once bending ceases and the metallic bonds are re-established, most of the residual strain is permanently retained, even after external force removal (P=0), except for a minor recoverable elastic component. The tendon-driven retraction (T>0) restores the fingers to their extended state. Pneumatic actuation ensures uniform stress distribution, extending the metal layer's fatigue life, while the metal layer, also as a constant layer, enhances bending stability. The outer chamber design further promotes uniform bending during tendon driven retraction, allowing the finger to consistently return to the same initial configuration. This improves the long term reliability of the gripper while ensuring that the metal layer maintains a single constant curvature arc throughout bending. Since the soft fingers are fabricated in a curved shape, the actuation to the neutral 0 °position establishes a pre-stretched layer, which mitigates the recoverable slip-induced deformation (elastic strain). As shown in Fig. 1C, the TSPM features a lead-screw-inspired mechanism. The screw assembly consists of two components fastened together using bolts. Tendons are secured to the screw via aluminum sleeves and are actuated through a centralized tendon routing system embedded between the two components. The tendon routing configuration and dimensions of the key structural components are illustrated in Fig. S1. During inflation, the motor reverses (T<0), releasing tendon tension while simultaneously driving the screw (external thread). The screw's frictional force induces the flanged nut (internal thread) to rotate, but its motion is constrained by a linear guide, thus converting rotation into linear translation of the paw pad. Consequently, a single motor-driven motion simultaneously adjusts the tendon tension and actuates the extension and retraction of the paw pad, enabling coordinated finger motion and top-side support.

Inspired by the paw pads of mammals such as cats and dogs, which primarily function to absorb impact forces, protect skeletal joints, provide friction and stability, and sense vibrations for rapid prey response, we designed and fabricated a biomimetic paw pad (mechanical characteristics shown in Fig. S2) embedded with a flexible force-sensitive resistor. During grasping, the paw pad deforms upon contact with the target object, offering compliant support from above and enhancing the gripper's adaptability and grasping stability. Although some bending angle is sacrificed due to the presence of the paw pad, the structure maintains omnidirectional support even after the elastic recovery of the paw pad and the embedded metal layer (Movie S2). This configuration effectively protects both the TSPM structure and the grasped object as well. In addition, integrated sensors enable the gripper to estimate the required gripping force (Movie S3) and facilitate rapid grasping (Movie S4).

## 2.2. Static Modeling of the Mechanism

To gain deeper insight into the CSSPD mechanism, we developed a static model to elucidate its underlying physical principles and optimize the kirigami structure. Specifically, we investigated how the plastic phase influences the continuous stability of the system. The expansion process is modeled under the quasi-static assumption. Owing to the material damping of the hyperelastic elastomer and the mechanical constraint imposed by the embedded metal sheet, the actuator deforms at a relatively low strain rate. Therefore, inertial effects and variations in kinetic energy are neglected, and the actuator is assumed to remain in mechanical equilibrium throughout the deformation process.

To describe the quasi-static equilibrium of the actuator, a total potential energy function $\Pi(\theta)$ is formulated by accounting for the elastic deformation of the soft elastomer, the pre-stretching layer, and the embedded metal sheet, as well as the plastic dissipation associated with the irreversible deformation of the metal sheet and the work performed by the pressurized air. Specifically, $U_s(\theta)$ represents the strain energy stored in the soft elastomer, $U_{PL}(\theta)$ denotes the elastic energy associated with the pre-stretching layer, $W_e$ is the recoverable elastic strain energy stored in the metal sheet, and $W_p$ represents the accumulated plastic work. The work performed by the internal pressure $P$ during the chamber volume change is expressed as $P\Delta V(\theta)$, where $\Delta V(\theta)$ is the change in chamber volume relative to the reference configuration. The generalized energy functional is consequently defined as

$$\Pi(\theta) = U_s(\theta) + U_{PL}(\theta) + W_p(\theta) + W_e(\theta) - P\Delta V(\theta) \tag{1}$$

Under the quasi-static assumption, the equilibrium bending angle is determined by the stationarity condition of the generalized energy functional with respect to $\theta$

$$\frac{\partial \Pi(\theta)}{\partial \theta} = 0 \tag{2}$$

Eq. (2) establishes the governing equilibrium condition for determining the actuator's bending angle under a prescribed internal pressure.

As shown in Fig. 2A, under PCC kinematics, each segment of the metal layer is assumed to undergo approximately uniform curvature. Since the metal ligament experiences small strain and the cross-section is assumed to remain plane and normal to the neutral axis, Euler–Bernoulli beam theory is adopted to describe the local bending strain. During deformation, metal can be divided into two regions: the elastic region and the plastic region. In the coordinate frame of Fig. 2B, the dividing line is $|y| = y_{\mathrm{p}}$.

To account for the nonlinear elastoplastic response and the Bauschinger effect during initial loading and unloading, a 1D elastoplastic constitutive model with Prager's linear kinematic hardening is formulated. The total strain increment $\mathrm{d}\varepsilon$ is decomposed into elastic and plastic components:

$$\mathrm{d}\varepsilon = \mathrm{d}\varepsilon_e + \mathrm{d}\varepsilon_p \tag{3}$$

The yield surface is defined by the classical yield function:

$$f(\sigma) = |\sigma - \alpha| - \sigma_Y \leq 0 \tag{4}$$

where $\sigma$ is the nominal normal stress, $\sigma_{\mathrm{Y}}$ is the initial yield strength, and α represents the back-stress capturing the kinematic translation of the yield surface (Bauschinger effect). According to Prager's kinematic hardening rule, the evolution of back-stress is proportional to the plastic strain increment:

$$\mathrm{d}\alpha = H_{\mathrm{kin}}\mathrm{d}\varepsilon_p \tag{5}$$

where $H_{\mathrm{kin}}$ is the constant kinematic hardening modulus of the material, determined from the experimentally measured plastic tangent modulus $E_t$ according to $H_{\mathrm{kin}} = EE_t/(E - E_t)$, where $E$ is Young's modulus. Under repeated actuation cycles (Fig. S3), the structural material transitions into an elastic shakedown state (or steady cyclic saturation). Once pre-conditioned beyond initial training cycles, additional irreversible plastic strain increments vanish ($d\varepsilon_p \to 0$). Consequently, the back-stress evolution ceases ($d\alpha = 0$), freezing the back-stress at a residual value $\alpha_{res}$. At this steady state, the effective yielding stress for subsequent loading cycles stabilizes as a constant:

$$\sigma_{Y,eff} = \sigma_Y + \alpha_{res} = \mathrm{constant} \tag{6}$$

To accurately capture the localized nonlinear plastic hardening in this steady state, the experimental stress–strain data during plastic deformation (Fig. S3) are fitted. The elastoplastic stress distribution across the beam cross-section can be expressed as:

$$\sigma(y) = \begin{cases} E\varepsilon_e = E\theta y/h, & |y| < y_{\mathrm{p}} \\ \sigma_y + E_t\Delta\varepsilon_p, & |y| > y_{\mathrm{p}} \end{cases} \tag{7}$$

$$\Delta\varepsilon_p = \varepsilon_{\mathrm{zz}} - \varepsilon_{\mathrm{Y}} \tag{8}$$

where $\Delta\varepsilon_p$ is the plastic strain.

Assuming the central axis of the beam does not change length, the strain in the beam is given by

$$\varepsilon_{\mathrm{zz}} = \frac{\theta y}{h} \tag{9}$$

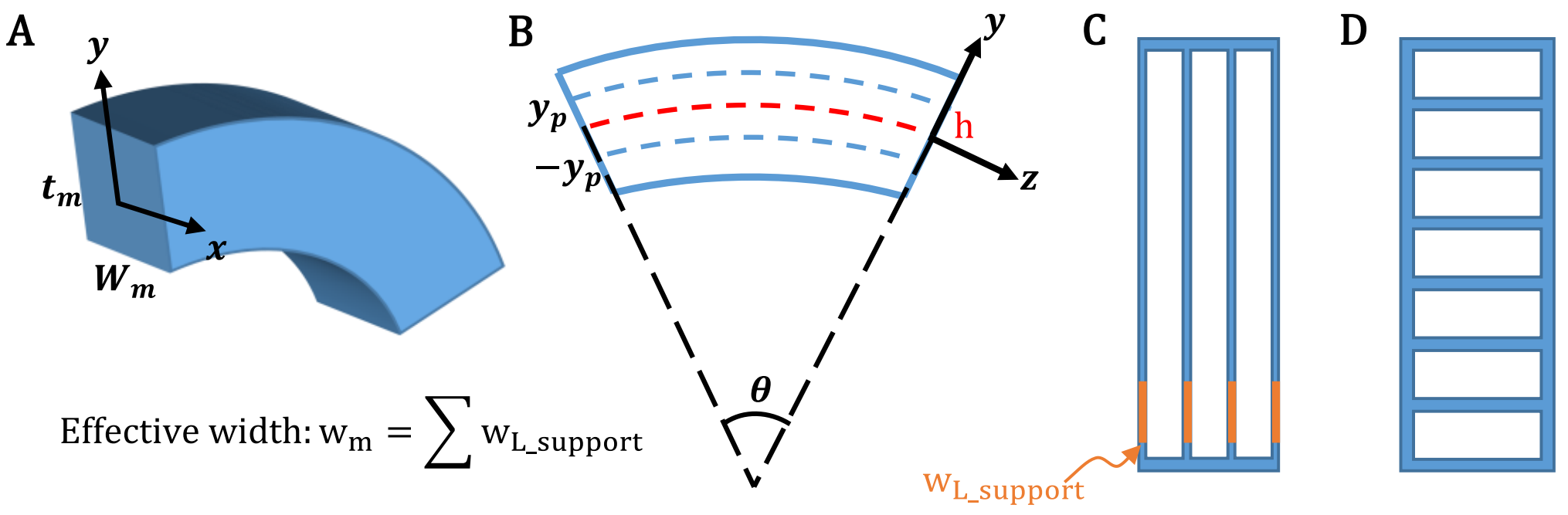

**Fig. 2. Configuration of metal layer.** (A) A cross junction of effective width $w_m$ and thickness $t_m$, with local coordinates x and y. (B) A cut view of the same cross-junction. h represents the length of the constant layer. Elastic stresses occur for $|y| < y_\mathrm{p}$, while plastic stresses occur for $|y| > y_\mathrm{p}$. (C) Long support configuration. $w_\mathrm{L_Support}$ is the width of each long support. (D) Short support configuration.

The moment of the metal layer also comes from the elastic and plastic regions.

$$\begin{cases} M_\mathrm{m}(\theta) = \int_A y\sigma(x,\theta)\,\mathrm{d}A \\ M_\mathrm{m}(\theta) = M_\mathrm{e}(\theta) + M_\mathrm{p}(\theta) \end{cases} \tag{10}$$

Therefore, the moment of the elastic and plastic regions can be calculated by

$$\begin{cases} M_\mathrm{e}(\theta) = \int_{-\frac{\mathrm{w_m}}{2}}^{\frac{w_\mathrm{m}}{2}} \int_{-y_\mathrm{p}}^{y_\mathrm{p}} y\sigma_e\,\mathrm{d}y\,\mathrm{d}x \\ M_\mathrm{p}(\theta) = 2\int_{-\frac{w_\mathrm{m}}{2}}^{\frac{\mathrm{w_m}}{2}} \int_{y_\mathrm{p}}^{\frac{t_m}{2}} y\sigma_p\,\mathrm{d}y\,\mathrm{d}x \end{cases} \tag{11}$$

$$y_\mathrm{p} = \frac{h\sigma_\mathrm{yield}}{E\theta} = \frac{hE\varepsilon_\mathrm{Y}}{E\theta} \tag{12}$$

where $\mathrm{w_m}$ is the effective width of the metal layer and $\mathrm{t_m}$ is the thickness of the metal layer.

The work of the elastic and plastic regions are obtained by integrating the bending moments with respect to the bending angle:

$$\begin{cases} W_\mathrm{e}(\theta) = \int M_\mathrm{e}(\theta)\,\mathrm{d}\theta \\ W_\mathrm{p}(\theta) = \int M_\mathrm{p}(\theta)\,\mathrm{d}\theta \end{cases} \tag{13}$$

From Eq. (11), it can be seen that the region of $|y_\mathrm{p}|$ (elastic deformation) is related to the bending angle. The thickness $t_\mathrm{m}$ influences the region of plastic deformation, which increases with thickness. And greater thickness means a larger driving force is required to actuate the metal layer. The derived analytical model shows that the mechanical response of the Kirigami layer depends explicitly on its effective length and width, whereas the detailed cutout geometry does not enter the governing equations independently. Therefore, other geometric configurations were not considered, except for the long-support and short-support designs, as illustrated in Fig. 2, C and D. We adopted a combination of both configurations, primarily selecting the long support structure with a few short supports to ensure reliability. The short support rods are used for structural stability and force transmission. Since the short support sections constitute a very small portion of the entire metal layer, we omit them in the mathematical model for simplification. Therefore, the effective width of the metal layer $\mathrm{w_m}$ equals the sum of all the long supports' widths.

Next, we mathematically model the soft region. In this paper, the 1st Yeoh hyperelastic model, which is equivalent to the Neo Hookean model, is used, where the energy density function $u_c$ can be written as

$$\begin{cases} u_c = C_{10}(I_1 - 3) \\ I_1 = {\lambda_1}^2 + {\lambda_2}^2 + {\lambda_3}^2 \\ \lambda_1\lambda_2\lambda_3 = 1 \end{cases} \tag{14}$$

where $C_{10}$ is the initial shear modulus. We assume the material is incompressible. $I_1$ is the first deviatoric strain invariant. $\lambda_1$, $\lambda_2$ and $\lambda_3$ denote the axial, radial, and circumference stretch rate, respectively.

For the expansion part, which is equi-biaxial stretch, the longitudinal stretch $\lambda_1$ equals the latitudinal stretch $\lambda_2$. Let $\lambda_1 = \lambda_{s1}$, the stretch in wall thickness $\lambda_3 = 1/(\lambda_{s1}^2)$ due to the incompressibility constraint. Hence,

$$u_e(\lambda) = C_{10}(2{\lambda_{s1}}^2 + 1/{\lambda_{s1}}^4 - 3) \tag{15}$$

For the pre-stretched part, the length in the width direction of the actuator is approximately unchanged, so $\lambda_2 = 1$. Let $\lambda_1 = \lambda_{s2}$, $\lambda_3 = 1/\lambda_{s2}$. Therefore,

$$u_c(\lambda) = C_{10}(\lambda_{s2} - 1/\lambda_{s2})^2 \tag{16}$$

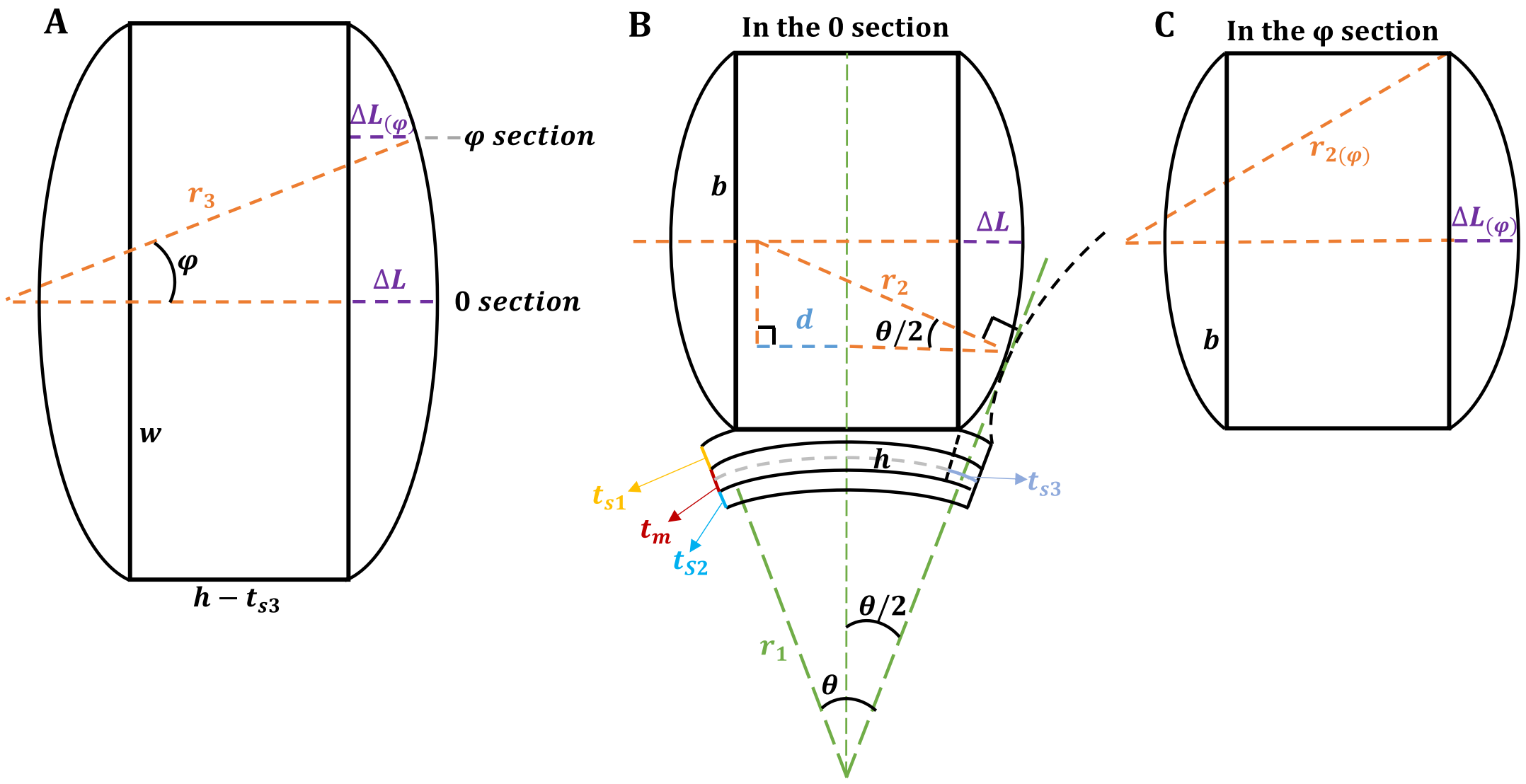

**Fig. 3. Configuration of soft finger.** (A) Top view of a single segment. (B) Front view single segment in the 0 section. (C) Front view single segment in the φ section. Where h and w represent the height and width of the segment. $t_m$ represents the thickness of metal. The soft finger has n semi-cylindrical segments. The spacing between different segments in the constant layer is $t_{s3}$. $t_{s1}$ and $t_{s2}$ represent the thickness of the rectangle parts on the bottom, which are also the pre-stretching parts.

At the initial stage of the collision, the contact point is located at the center, specifically at the point of maximum displacement $\Delta L$ of the expansion layer (in the 0 section). Due to the relatively high hardness of the TPU material, it will not compress when in contact with other expansion layers after inflation. It consistently maintains point contact, causing the contact point to shift toward the constant layer (the central axis of the metal layer). The radius of curvature of the expansion layer is $r_2$, which is perpendicular to the bending radius at the contact point. As shown in Fig. 3, since the bottom of the expansion part is very close to the constant layer, we assume that $h - t_{s3}$ is a constant during the bending.

Based on the triangular relationship, $\mathrm{r}_1$, $\mathrm{r}_2$ and $\mathrm{r}_3$ can be described by

$$\begin{cases} r_1 = \dfrac{h}{\theta} \\ r_2 = \dfrac{b^2}{8\Delta L} + \dfrac{\Delta L}{2} \\ r_3 = \dfrac{w^2}{8\Delta L} + \dfrac{\Delta L}{2} \end{cases} \tag{17}$$

Through trigonometric relationships, we can get the relationship between θ and ΔL.

$$r_2 \cos\frac{\theta}{2} - d = \left(r_1 + \frac{t_m}{2} + t_{s1} + \frac{b}{2} - r_2 \sin\frac{\theta}{2}\right) \tan\frac{\theta}{2} \tag{18}$$

$$d = r_2 - \Delta L - \left(\frac{h}{2} - \frac{t_{s3}}{2}\right) \tag{19}$$

To determine the function $\Delta L = f(\theta, t_{\mathrm{m}})$, numerical values were substituted as follows: $b$=12, $h$=7, $t_{\mathrm{s}}$=1, $t_{\mathrm{s1}}$=2.8, $t_{\mathrm{s2}}$=1, $t_{\mathrm{s3}}$=1, $w$=25 based on Eq. (17) to (19). Using the MATLAB Curve Fitting Toolbox, the fitting yielded Eq. (20) with an SSE of 0.6458 and an RMSE of 0.0080.

$$\Delta L = 4.34\theta + 0.04583 t_{\mathrm{m}} + 0.4617 \tag{20}$$

The maximum deformation $\Delta L_{(\varphi)}$ in the expansion direction on the $\varphi$ section is

$$\Delta L_{(\varphi)} = \Delta L - (r_3 - r_3 cos\varphi) \tag{21}$$

Therefore, on the cross section of any angle φ

$$r_{2(\varphi)} = \frac{b^2}{8\Delta L_{(\varphi)}} + \frac{\Delta L_{(\varphi)}}{2} \tag{22}$$

$$\lambda_{s1} = \frac{2r_{2(\varphi)}}{b} \cdot \arcsin\frac{b}{2r_{2(\varphi)}} \tag{23}$$

$$U_{\mathrm{s1}}(\theta) = 2\int_{V_{\mathrm{s1}}} u_{\mathrm{e}}(\lambda_{s1})\,\mathrm{d}V_{\mathrm{s1}} \tag{24}$$

$$\Delta V(\theta) = 4\int_0^{\arcsin\frac{w}{2r_3}} \left(\frac{r_{2(\varphi)}\lambda_{s1} b}{2} - \frac{\left(r_{2(\varphi)} - \Delta L_{(\varphi)}\right) b}{2}\right) r_3 \cos\varphi\,\mathrm{d}\varphi \tag{25}$$

Due to the complexity of the integrals in Eq. (24) and (25), their numerical solutions were first evaluated in MATLAB. A total of 10,000 sample points were generated over $\theta \in [0, \pi/n]$, and $t_{\mathrm{m}} \in [0.5, 1.5]$ the corresponding theoretical values of $U_{\mathrm{s1}}(\theta)$ and $\Delta V(\theta)$ were calculated from the Eq. (24) and (25). Polynomial surface fitting was then performed on these numerically calculated theoretical data with respect to $\theta$ and $t_{\mathrm{m}}$, yielding simplified explicit expressions for $U_{s1}(\theta)$ and $\Delta V(\theta)$. The resulting fitting surfaces are shown in Fig. S4, and the corresponding polynomial expressions are given in Eq. (26).

$$\begin{cases} U_{s1}(\theta) = 821.6\theta^3 + 59.08\theta^2 + 9.162\theta t_m \\ \Delta V(\theta) = 1188\theta + 12.55t_m + 121.1 \end{cases} \tag{26}$$

For the calculation of the pre-stretched layer, in the design, we set the initial bending angle of a single segment to $\theta_{\mathrm{PS}}$, then the stretching rate $\lambda_{s2}$ at the bending angle $\theta$ is:

$$\lambda_{s2} = (h + x\theta)/(h + x\theta_{PS}) \tag{27}$$

$$U_{s2}(\theta) = \int_{V_{s2}} u_c(\lambda_{s2})\,\mathrm{d}V_{s2} \tag{28}$$

When the tendon is pulled back, the gaps between each segment of the finger allow the structure to be pulled to an angle of $\theta_{\mathrm{r}} = -\frac{9\pi}{180\mathrm{n}}$, where n is the number of segments (set to 9 in our design). Due to the elastic recovery following plastic deformation of the metal layer and the torque generated by the pre-stretching layer, the structure eventually stabilizes at an angle $\theta_0$. Since the system exhibits history dependency, the final angle is influenced by the previous bending angle. To simplify the model, only the metal bending resistance against $M_{\mathrm{PL}}$ is considered in this analysis.

$$M_{PL}(\theta_0) + M_m(\theta_0 - \theta_r) = 0 \tag{29}$$

When $\theta = \theta_0$ is taken as the reference position, the value of $U_{\mathrm{PL}}(\theta)$ (the rectangle part on the bottom, which is the pre-stretching layer) should be the difference between $U_{\mathrm{s2}}(\theta)$ and $U_{\mathrm{s2}}(\theta_0)$ at the reference position.

$$U_{\mathrm{PL}}(\theta) = U_{\mathrm{s2}}(\theta) - U_{\mathrm{s2}}(\theta_0) \tag{30}$$

According to the above equations, the relationship between the bending moment, strain energy, and bending angle during pressurization can be established. After the pneumatic pressure is released, the segments partially recover their bending angles and reach a new equilibrium at $\theta_{\mathrm{f}}$, where the moment generated by the pre-stretching layer balances the restoring moment associated with the metal layer.

Predicting elastic recovery (springback) remains a longstanding and challenging research topic. Previous experimental studies have reported that the unloading modulus of metallic materials can decrease with increasing plastic deformation, potentially influencing the accuracy of elastic recovery predictions [53, 54]. In the present system, the discrepancy between the predicted and experimentally measured recovery angles may also arise from several coupled effects, including plastic-strain-dependent modulus degradation, changes in the effective arc length of the deformation region with increasing bending angle, and thickness thinning of the metal layer. To account for these effects in a simplified manner without introducing additional constitutive and geometric complexity, an effective unloading modulus $E_{um}$ is introduced as

$$E_{um} = (1 - d_E)E \tag{31}$$

where $E$ is the initial Young's modulus and $d_E$ is the unloading-modulus degradation factor. In this study, $d_E = 0.3$ is adopted as a preliminary modeling assumption to investigate the influence of reduced unloading stiffness on the predicted recovery angle. The modified modulus is used only in the rebound equation, while the original constitutive model is retained for the loading process.

Assuming that the recovery process can be approximated as elastic unloading, the rebound equilibrium equation is modified as

$$M_{\mathrm{PL}}(\theta_{\mathrm{f}}) + \frac{(\theta - \theta_{\mathrm{f}})E_{um}I}{h} = M_{\mathrm{m}}(\theta) \tag{32}$$

Here, $M_{\mathrm{m}}(\theta)$ is obtained from Eq. (10), and $M_{\mathrm{PL}}(\theta_{\mathrm{f}})$ is calculated from the derivative of the pre-stretching-layer strain energy with respect to the final bending angle. The unloading-modulus correction is applied only to the elastic unloading term, while the original loading constitutive model is retained. The assumed degradation factor is evaluated by comparing the predicted recovery angles with experimental measurements.

$$M_{\mathrm{PL}}(\theta_{\mathrm{f}}) = \frac{\mathrm{d}U_{\mathrm{PL}}(\theta_{\mathrm{f}})}{\mathrm{d}\theta_{\mathrm{f}}} \tag{33}$$

Since the bending moment $M_{\mathrm{m}}$ remains the same whether the structure bends forward from an initial angle of $\theta_0$ to $\theta_{\mathrm{f}}$ or reverse its bending direction, the required tendon-driven torque can be determined using Eq. (34). Based on the optimization results of the metal layer presented in Text S1, a 3D-printed servo horn with a radius of 21 mm was ultimately selected to satisfy the design requirements.

$$M_{\mathrm{t}} = M_{\mathrm{m}}(\theta_{\mathrm{f}} - \theta_0) - M_{\mathrm{PL}}(\theta_0) \tag{34}$$

Then the final bending angle $\theta_{\mathrm{SF}}$ and recovery angle $\theta_{\mathrm{SR}}$ of the soft finger can be obtained as

$$\begin{cases} \theta_{\mathrm{SF}} = n\theta_{\mathrm{f}} \\ \theta_{\mathrm{SR}} = n(\theta_{\mathrm{f}} - \theta) \end{cases} \tag{35}$$

The bending stiffness of the segment $K$ can be expressed as

$$K = \frac{\mathrm{d}M_{\mathrm{PL}}(\theta_{\mathrm{f}})}{\mathrm{d}\kappa_f} + EI \tag{36}$$

where $\kappa_{\mathrm{f}} = \theta_{\mathrm{f}}/n$, is the curvature corresponding to $\theta_{\mathrm{f}}$:

Finally, the bending stiffness of the whole soft finger $K_{\mathrm{SF}}$ can be expressed as

$$K_{\mathrm{SF}} = K/n \tag{37}$$

## 2.3. Characterization of the Mechanism

The two most critical factors governing the CSSPD mechanism are the bending stiffness and the elastic recovery (rebound) of the metal layer. Achieving passive stability solely through the material’s intrinsic properties, without external structures or forces, inevitably results in rebound. This is because the material needs external forces to deform and reach mechanical equilibrium. Upon removing these external forces, a rebound occurs, leading to a new equilibrium. The stable behavior of the CSSPD mechanism arises as, during the rebound process, the elastic recovery ceases, and then the metal layer reaches a new balance, which generates a resisting force in the opposite direction of the external load. In our design, this rebound is mitigated by incorporating a pre-stretched layer, and further neutralized through the TSPM. By allowing the finger to restore its angle in advance, the elastic recovery is counteracted, thereby enhancing the stability of the grasp. To qualitatively assess the influence of various pre-stretched layer sizes and metal layer kirigami structure on the CSSPD mechanism, experiments and models for specific design variables were compared, as shown in Fig. 4. The thickness of the pre-stretched layer (Fig. 4A) influences the variation of the bending moment with respect to the bending angle. Because the resisting moment provided by the metal layer is primarily determined by its elastic modulus and geometric configuration, a steeper decrease in the net bending moment with increasing bending angle indicates a greater reduction in the effective restoring stiffness and, consequently, reduced structural stability. Therefore, a thinner pre-stretched layer is preferable from the perspective of maintaining structural stability. However, an excessively small thickness may compromise the sealing and structural integrity of the pneumatic chamber. Considering this trade-off, a thickness of 2.8 mm was selected. The initial angle of the pre-stretched layer $\theta_{\mathrm{PS}}$ was also investigated (Fig. 4B). When the bending angle is smaller than $\theta_{\mathrm{PS}}$, the direction of the moment generated by the pre-stretched layer opposes that produced by the elastic recovery of the metal layer, resulting in a smaller rebound of the metal layer. Changing $\theta_{\mathrm{PS}}$ does not substantially alter the decreasing trend of the bending moment with increasing bending angle. However, an excessively small $\theta_{\mathrm{PS}}$ results in a large angular recovery ($\Delta\theta$), at the maximum bending angle, whereas an excessively large $\theta_{\mathrm{PS}}$ makes it difficult for the CSSPD finger to return to and stabilize at the 0 °configuration during retraction. Therefore, $\theta_{\mathrm{PS}}$ was selected by considering the required driving pressure, angular recovery, and available working range. Based on this trade-off, an initial angle of 22 °for each section of the pre-stretched layer was selected for the final design. Due to thermal shrinkage of the TPU during the 3D printing process, the fabricated sections exhibited a reduced pre bending angle of 18 °. With a total of nine sections, this resulted in an overall initial angle of 162 °for the finger.

The metal sheets used in the experiment are shown in Fig. S5. The mathematical model and experimental results for the bending stiffness of metal layers ($K_{\mathrm{m}}$) with varying widths ($w_{\mathrm{m}}$) as a function of thickness ($t_{\mathrm{m}}$) (Fig. 4C) fit well. However, there is some discrepancy between the model and experiments regarding the relationship between bending angle and pressure (Fig. 5, A-C) and the relationship between $\Delta\theta$ and bending angle for different thicknesses (Fig. 5, D-F). The experiments were repeated three times, and the averaged values are presented in the figures. The main sources of error are as follows:

1. The constraint layer was assumed to be located at the mid-plane of the metal layer, and the pre-strain distribution was assumed to be uniform.

2. The effects of material hardening and the Bauschinger effect were neglected in the analytical model.

3. A simplified rigid-plastic material model was adopted, and the plastic deformation of the kirigami metal layer was assumed to be homogeneous.

The experimental repeatability error was minimal, primarily due to the design of the finger's pneumatic chamber, which ensured a consistent initial bending angle each time the tendon was fully actuated. The main source of uncertainty arose from the measurement of the bending angle, with an estimated error of approximately ±1 degree. This error was not visually apparent in Fig. 5. and therefore not indicated. However, since the bending angle error significantly affects the stiffness calculation, error bars were included in Fig. 4C to reflect this uncertainty.

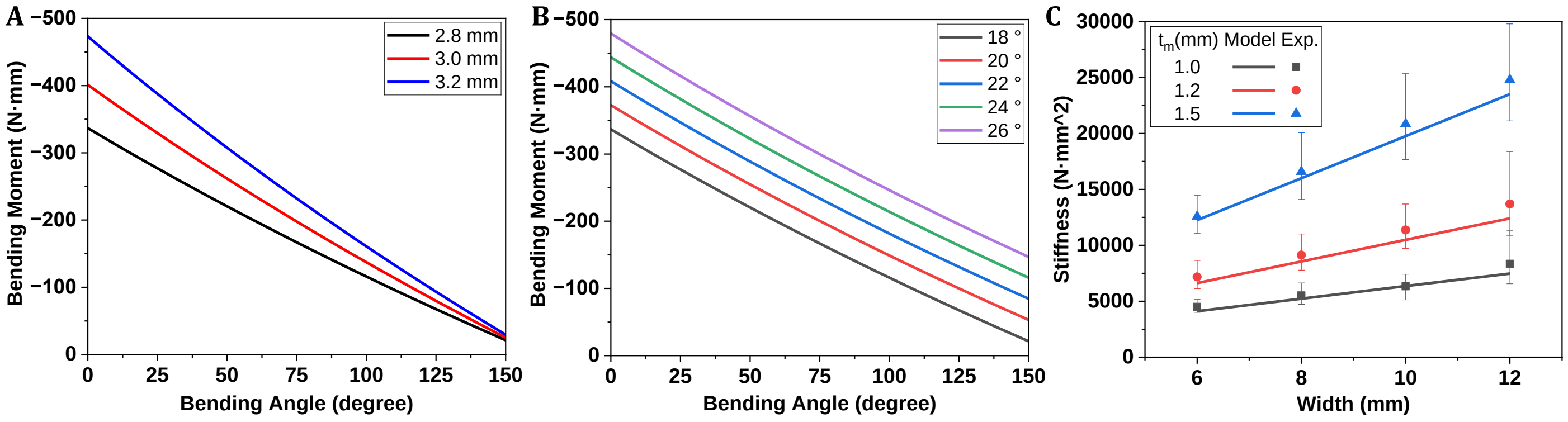

**Fig. 4. Characterizations of the pre-stretched layer and the stiffness of CSSPD.** Bending moment of the pre-stretched layer as a function of the bending angle for (A) different thicknesses with 18 ° initial angle of each section, and (B) different initial angles of each section with 1.2 mm thickness, respectively. (C) Bending stiffness as a function of different widths ($W_m$) for different thicknesses ($t_m$).

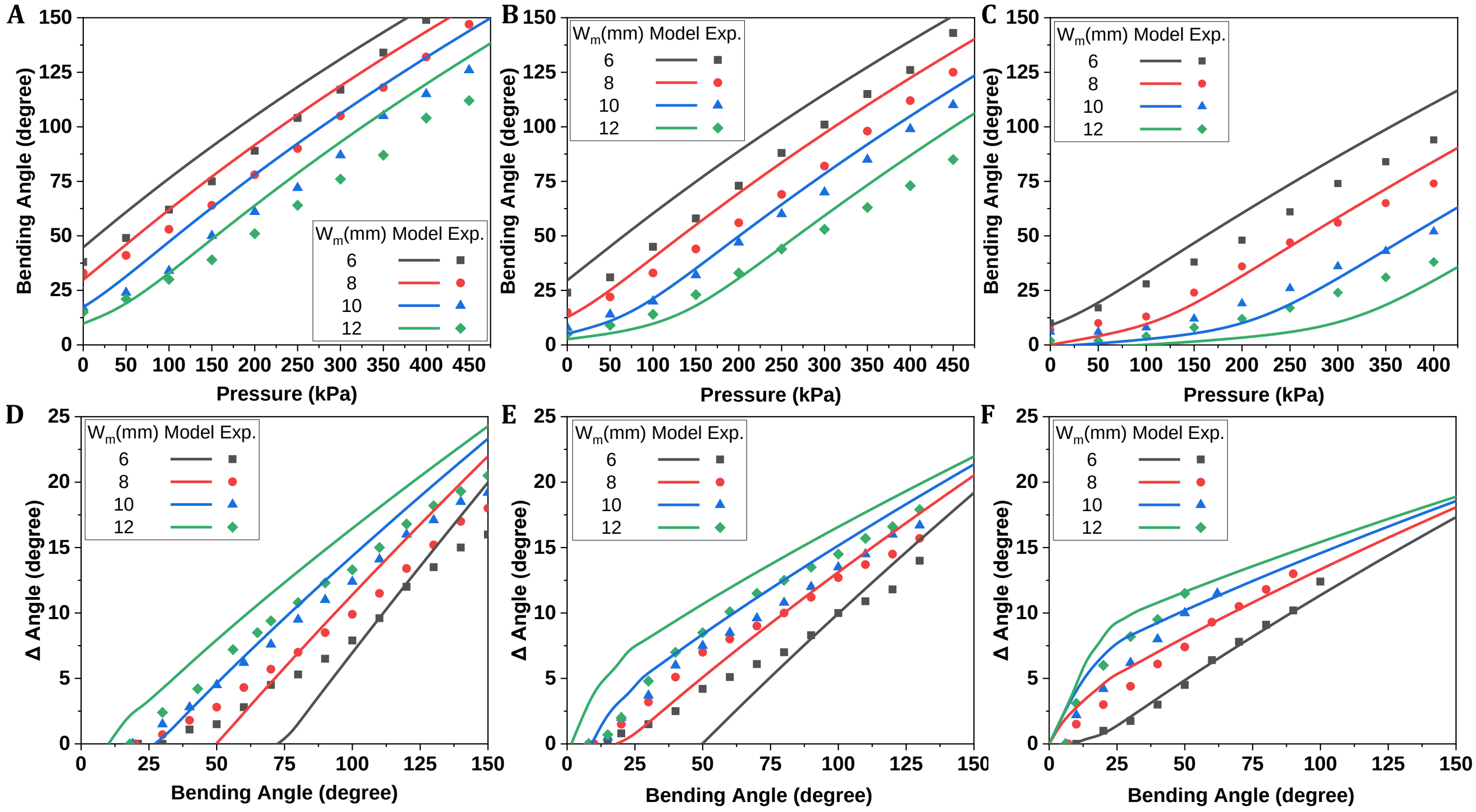

**Fig. 5. Characterizations of the CSSPD through modeling and experiments.** Modeling results illustrating the relationship between air pressure and bending angle for different thicknesses of metal layer (A) 1 mm. (B) 1.2 mm. (C) 1.5 mm. Corresponding comparisons between experimental and modeling results for bending angle and recovery angle (Δangle) with respect to different widths, respectively, are shown for the same thickness in (D) 1 mm, (E) 1.2 mm, (F) 1.5 mm. Solid lines represent model predictions, while experimental data are denoted by symbols.

The results showed that larger $t_\mathrm{m}$ and smaller $w_\mathrm{m}$ reduce rebound. Based on the optimization framework established in Text S1, we ultimately chose $t_\mathrm{m}$ of 1.2 mm and $w_\mathrm{m}$ of 6 mm. A thicker layer would require higher driving air pressure, and excessively narrow width could lead to irregular bending in the width direction after multiple cycles, compromising the structural stability. Generally, for a given input air pressure, larger bending angles result in smaller output forces and lower stiffness. However, in our structure, the stiffness remains almost constant at any bending angle, primarily due to the elastic stiffness of the metal layer. This stiffness exists even without input air pressure and increases with the assistance of the driving pressure. The CSSPD mechanism demonstrates bidirectional stiffness and stability at all bending angles.

# 3. Experimental Results

This section demonstrates the static and dynamic stability of the proposed gripper through experiments. It then highlights its strong potential for diverse applications through demonstrations of versatile grasping, rapid automatic grasping, and energy free stable perching on a branch.

### 3.1. Static Stability Experiments

To evaluate the static stability of the CSSPD mechanism, we conducted experiments under three conditions by varying the presence of pneumatic actuation (0.3 MPa) and the embedded metal layer (Movie S5). In each case, the CSSPD was used to grasp a 150 g ball-shaped object with an internal iron core, and a vertical downward force was applied using a digital force gauge to measure the maximum load the gripper could resist in a static state.

The results (Fig. 6 A) show that the CSSPD with pneumatic actuation but without the metal layer sustained a maximum static load of 19 N. Conversely, without actuation but with the embedded metal layer, the structure still supported up to 16 N, indicating that the metal layer alone contributes significantly to load-bearing through its inherent stiffness. When both the metal layer and pneumatic actuation were present, the gripper achieved a maximum static load of 28 N (approximately the additive sum of the two individual contributions), demonstrating a synergistic enhancement in stability.

These findings demonstrate the excellent static stability of the CSSPD mechanism. At the same bending angle, the metal layer undergoes plastic deformation during actuation, resulting in a permanent residual curvature after the applied load is removed. This residual deformation helps maintain the finger configuration and provides passive mechanical resistance, thereby supplementing the pneumatic actuation force and enhancing the overall load-bearing capacity of the gripper.

### 3.2. Dynamic Stability Experiments

To evaluate the dynamic stability of the proposed gripper, we conducted vertical stability (Movie S6) and horizontal stability experiments (Movie S7). Impact testing is a common approach to evaluate system stability [49, 55]. Similar to the static stability experiments (external loading), the vertical stability experiment involved the gripper holding a 150 g object, which was dropped to impact a PVDF sensor. Upon impact, a pulsed acceleration generated an upward reaction force on the gripper, while the grasped object continued downward due to inertia. This experiment effectively simulates grasp stability during rapid upward motion. Results showed that under a pulsed acceleration of 400 $m/s^2$ the gripper successfully retained the object in all three conditions: (i) no metal layer with 0.3 MPa pressure, (ii) metal layer without actuated pressure, and (iii) metal layer with 0.3 MPa. Notably, the metal layer alone provided considerable stability without pneumatic assistance, and the combination with actuated pressure further enhanced performance.

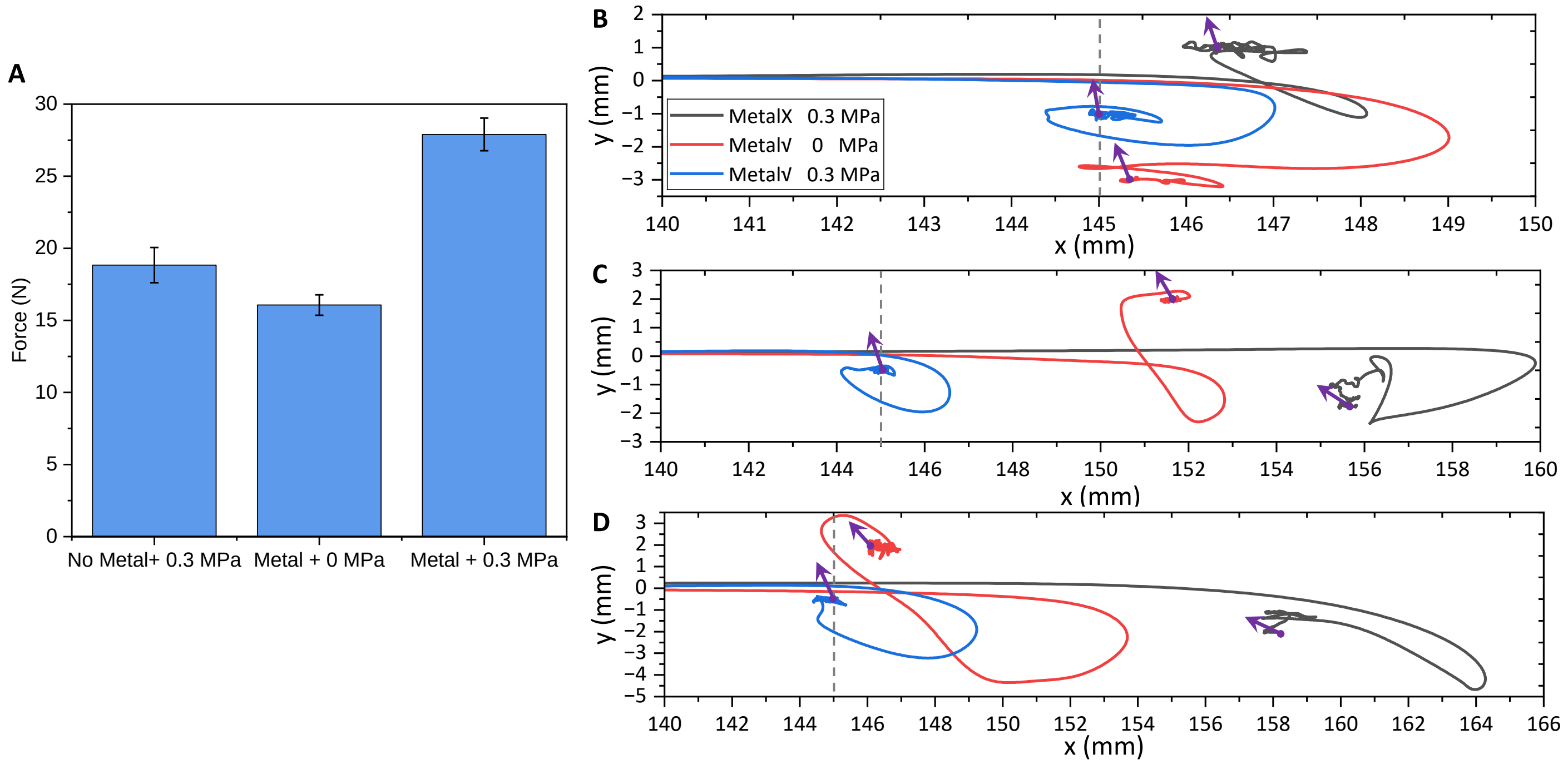

**Fig. 6. Experimental results of static and horizontal stability.** (A) Static stability experiments using a 150 g ball-shaped object with an internal iron core. A vertical downward force was applied using a digital force gauge to determine the maximum load that the gripper could withstand under static conditions. The maximum holding force was measured three times for each condition. (B–D) Qualitative results of horizontal stability experiments under three levels of pulsed acceleration with peak values of (B) 133 m/s$^2$, (C) 266 m/s$^2$, and (D) 400 m/s$^2$. The black, red, and blue solid lines represent the two-dimensional trajectories of the center of the grasped object under three conditions: no metal layer with 0.3 MPa pressure, metal layer without actuated pressure, and metal layer with 0.3 MPa pressure, respectively. Impact occurs when the 150 g grasped object reaches 145 mm in the *x*-direction, indicated by the gray dashed line. The small circles and arrows mark the final stabilized positions and angular deviations of the object after impact.

For the gripper, whether due to collisions or rapid motion, horizontal impacts are the dominant source of dynamic disturbance, making them more indicative of grasping stability. As shown in Fig. 6 B-D, under all three acceleration levels, some displacement occurred in the y-direction. The gripper with a metal layer showed improved stability in both x and y directions and reduced angular deviation compared to the configuration solely by 0.3 MPa pressure. When both the metal layer and 0.3 MPa pressure were applied, x-direction displacement was nearly eliminated, and both y-direction and angular deviations were significantly reduced, demonstrating the substantial stabilizing effect of the metal layer. Moreover, the impact experiment videos clearly show that the paw pad structure also contributes to enhanced vertical (y-direction) stability.

### 3.3. Grasping Experiments

The diversity of grasping is also a critical metric for evaluating gripper performance and warrants investigation. The proposed CSSPD gripper can grasp objects with a minimum diameter of 14 mm in the inflated state, which corresponds to the residual gap between the circumferentially arranged fingertips when they come into contact upon actuation. In the deflated state, due to recovery from the plastic deformation of the metal layer, the minimum stable graspable diameter increases to 17 mm. The TSPM structure provides a 13 mm extension range, enabling the paw pad to make contact with the object, thereby enhancing grasping stability. The gripper can reliably grasp and release objects of varying weights, hardness, and irregular shapes without requiring continuous pneumatic input (Movie S8). Each grasping trial was repeated three times, and all trials were successfully completed. As demonstrated in Fig. 7, it successfully holds items ranging from a lightweight 15 g corner bracket to a heavy ball with a load (equivalent to 1600 g). The gripper can also securely handle a wide variety of materials, from fragile items such as potato chip to rigid components like aluminum (AL) profile. It is capable of adapting to challenging objects, including wet water pouch, dry clothing, a thin CD, and a bunch of grapes.

### 3.4. Rapid Grasping Capability

Perception enables grippers to become intelligent, allowing for autonomous object detection and adaptive grasping. As shown in Fig. 8A, an experiment was conducted to demonstrate the proposed gripper's rapid response and stable grasping capability. In this test, a 150 g spherical object was dropped at approximately 4.5 m/s and impacted the center of the gripper, where the bioinspired paw pad is located. The pressure sensor embedded in the paw pad detected the impact force. Once the detected force exceeded a predefined threshold, the valve was triggered to open, allowing the fingers to rapidly inflate and grasp the object. With the pneumatic pressure set to 0.3 MPa, the system responded by triggering the paw pad and completed the grasp within approximately 0.5 seconds (Movie S4). The response time can be further reduced by using shorter air tubes and increasing the driving pressure.

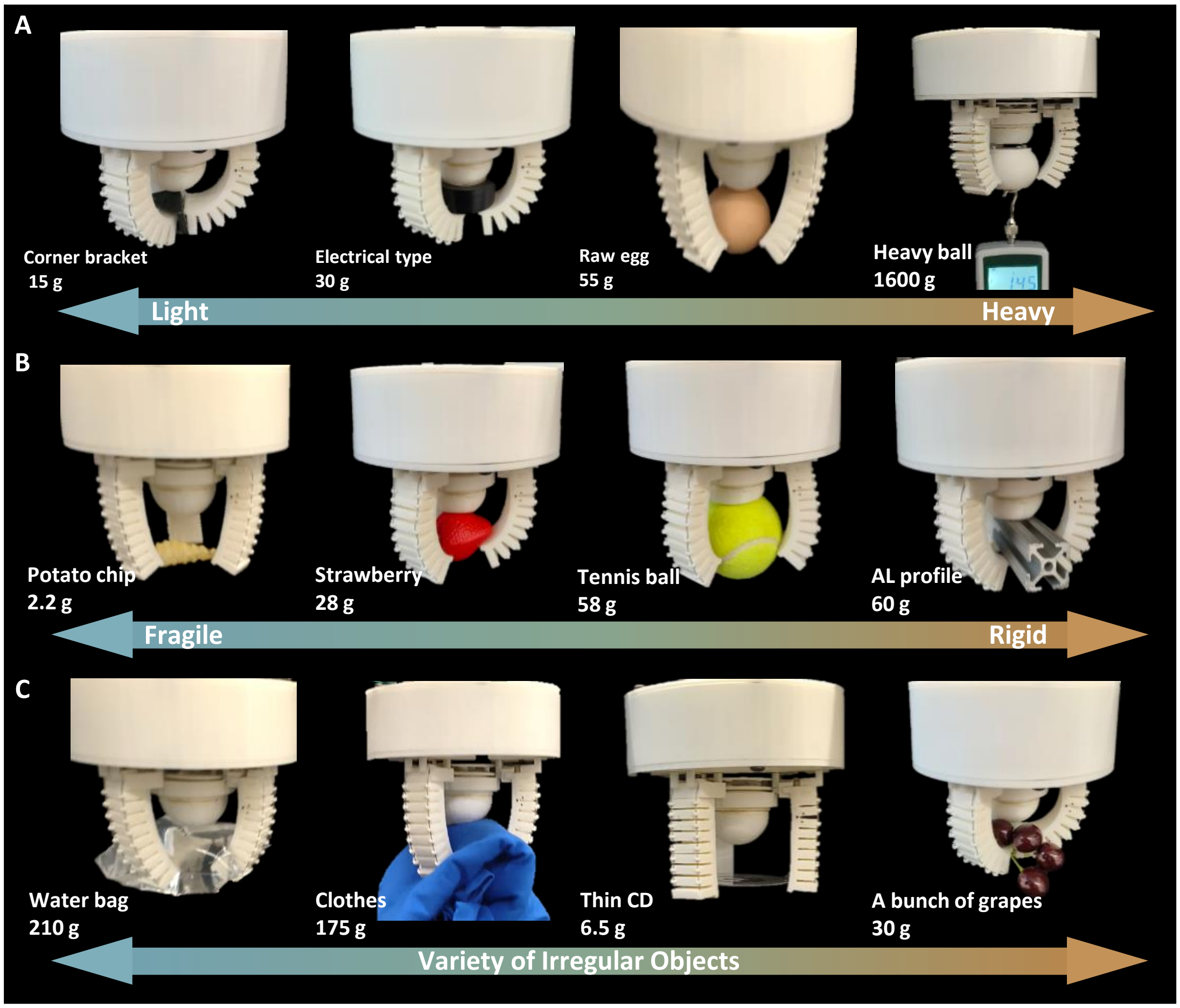

**Fig. 7. Demonstration of high load capacity and good shape adaptability of the gripper.** (A) Reliable grasping of objects ranging from lightweight to heavy. (B) Securely grasping objects from fragile to rigid. (C) Grasping a variety of challenging, irregular objects.

### 3.5. Energy-Free Stable Holding Capability

The ability to maintain a stable grasp at arbitrary bending angles without continuous energy input is essential for reliable long-term object holding. To demonstrate this feature, the proposed gripper was used as a mobile manipulator to grasp and perch on tree branches, as shown in Fig. 8B. For improved grasping of slender, rod-like branches, the gripper's original circumferential configuration was modified to a parallel configuration. This functionality could also be achieved through variable-structure [56], allowing the gripper to adapt its configuration according to different application scenarios in the future. In the demonstration, the gripper approached the branch and grasped it by actuation. It was then deflated to enter a passive holding state. During the deflation process, the gripper exhibited a slight rotational motion around the branch axis (Movie S9). This was mainly caused by the torque generated by the gripper’s own weight (660 g) on the rotational axis. Although this motion was subsequently counteracted by the elastic recovery of the paw pad, the changing friction conditions during the transition still resulted in minor rotational movement. After complete deflation, the gripper was able to perch stably on both horizontal and inclined branches, without requiring any energy input to maintain its position.

As the branch inclination increases toward the vertical direction, stable perching becomes more challenging because the gravitational component acting along the branch increases. For vertical branches, a diameter of 35 mm represents a limiting case for passive grasping with the pneumatic fingers. As the branch diameter increases, the fingers undergo a smaller elastic recovery angle after deflation, resulting in greater contact engagement and improved grasping stability. A branch diameter of approximately 50 mm corresponds to the spacing between the fingers in the parallel configuration. For branches larger than this spacing, tendon-driven actuation can further open the fingers beyond the neutral configuration, enabling the fingers to achieve a negative opening angle and mechanically interlock with the branch. This provides an additional mechanism for maintaining stable contact with larger branches.  In practical deployment, however, the stability of perching is affected by multiple factors, including platform vibration, the relative position and grasping angle of the fingers, the complex geometry of tree branches, interference from leaves, and the resulting frictional conditions. These factors are difficult to apply and control in a consistent and repeatable manner under our current experimental setup. Nevertheless, this energy-free holding capability suggests strong potential for integration with mobile platforms such as drones or underwater robots, where the gripper could remain securely attached to tree limbs or submerged rocks for extended durations, even under disturbances like wind or ocean currents. To further demonstrate its application in mobile platforms, the CSSPD fingers were arranged in a quadruped robot configuration capable of locomotion along a tube. A solar panel mounted on the robot can be deployed via shape memory alloy actuation (Fig. 8C). Locomotion along the tube is achieved through the reciprocating motion of pneumatic cylinders (Fig. 8D and Movie S10). This showcases a potential application in which the robot can perch on an object to recharge when the battery is low, and continuously monitor its surroundings if equipped with a camera, relying on solar energy as its power supply.

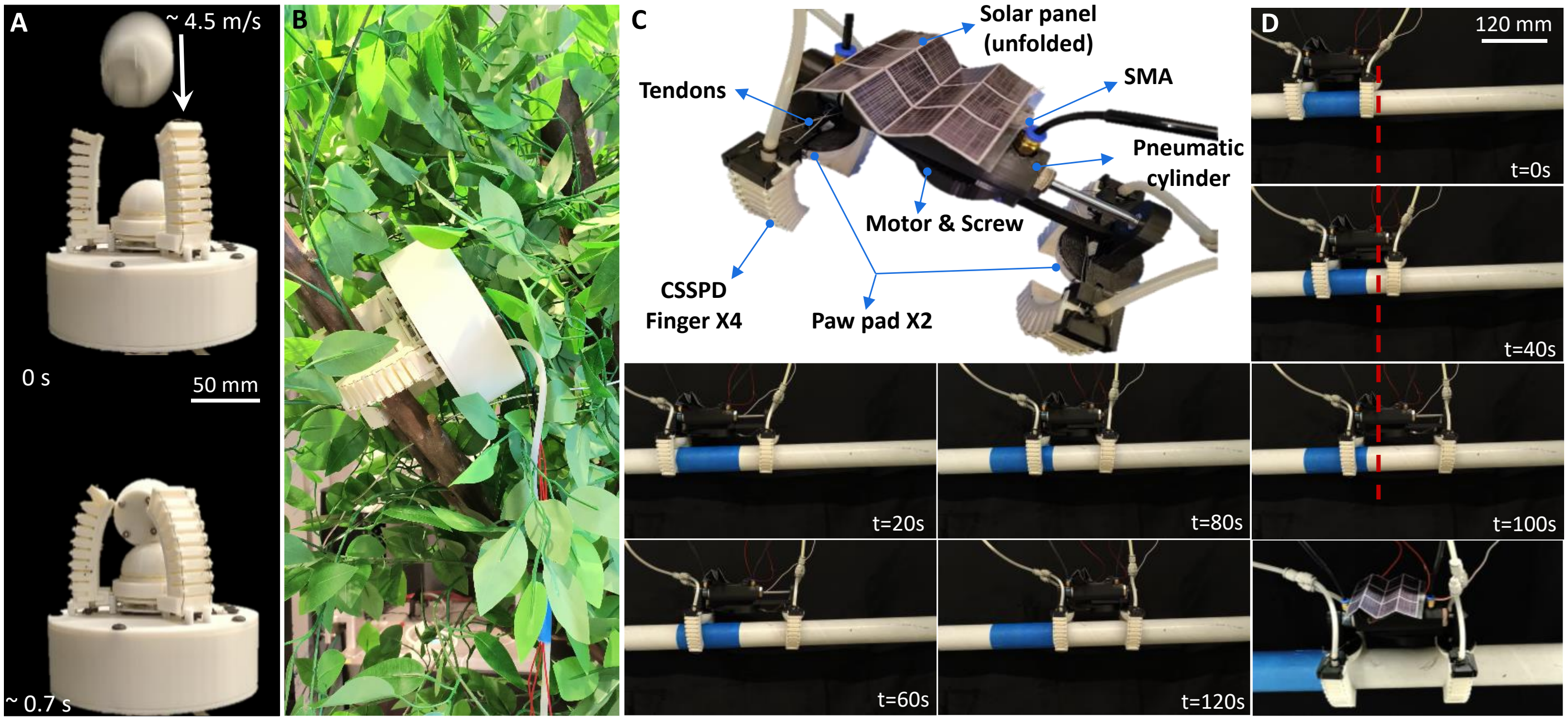

**Fig. 8. Demonstration of rapid grasping capability and tree grasping performance of the gripper.** (A) Rapid object grasping triggered by stress response. (B) Stable holding on a tree branch without continuous energy input. (C) Schematic of the quadruped robot structure based on the CSSPD. (D) Locomotion along a cylindrical surface.

# 4. Materials and Methods

In this section, the fabrication process and material characterization of the proposed gripper are presented, followed by the mechanical testing of the metal layer and the experimental setup, including the control system, hardware configuration.

## 4.1. Fabrication of Gripper

The rigid components were 3D printed using a high-resolution 3D printer (ESun PLA+HS, Bambu Lab P1P). The soft components (fingers and the soft paw pad) were fabricated using TPU material (NINJAFLEX TPU). Since the soft fingers are pneumatically actuated and require airtightness, the parameter settings in the slicing software (Bambu Studio) play a critical role in ensuring their performance. The key parameters are defined in Table S1. For the metal layer, we selected 1060 aluminum (GB/T 3190-1996, Southwest Aluminum Company), which was fabricated using CNC laser cutting. The chemical composition is provided in Table S2. The fatigue life analyze of metal layers is provided in Text S2.

## 4.2. Mechanical Testing

Tests were conducted under uniaxial loading on an Instron 5982 mechanical testing machine with a 100 kN axial load at a loading rate of 0.5 mm/s (Fig. S6). The mechanical data were analyzed with Origin. The Yeoh 1st hyperelastic model was employed to fit the parameters TPU, utilizing experimental data from the tensile testing. C10=3.2872MPa. The structural and material parameters are list in Table S3.

Due to the stress-strain behavior of 1060 aluminum, which initially varies with the number of times the material enters the plastic phase, it gradually stabilizes after several cycles. We conducted tensile testing on aluminum samples subjected to different numbers of bending cycles (0, 5, 25, 50, 75, 100) (Fig. S5). The experimental results indicate that the material reaches a near-stable state after 50 bending cycles. Based on these observations, the material parameters for 1060 aluminum were determined.

### 4.3. Experimental Setup

The control system is shown in Fig. S7**.** The data is collected by an Arduino microcontroller and subsequently processed on a laptop. Tendon actuation is achieved using a 0.4 mm diameter 304 stainless steel cable, driven by a high-torque digital servo (TIANKONGRC, 60 kg, SHENZHEN SKY STAR TECHNOLOGY CO., LTD., China). The servo operates at 6~8.4 V, provides a maximum stall torque of 60 kgf cm, and supports a rotation angle up to 270°. Pneumatic control of the soft fingers on the gripper is implemented using an electric proportional valve (AirTac 4V130C-06, CO, China), which modulates airflow from a pump supplying 0.3 MPa. A flexible force-sensitive resistor (FSR, RP-C18.3, sensing range: 20 g~6 kg, Interlink Electronics Inc., USA) was employed to detect contact forces. The FSR signal was conditioned through a voltage conversion module to generate analog signals readable by the microcontroller. For dynamic stability experiments, a polyvinylidene fluoride (PVDF) piezoelectric sensor (LDT0-028K, Vkinging Corporation, Shenzhen, China) was similarly integrated via a compatible voltage conversion module. PVDF is particularly suited for transient force detection due to its high sensitivity, rapid response time, and wide dynamic range.

As shown in Fig. S8, the experimental setup consisted of a linear rail rigidly fixed at both ends to a workbench. A PVDF sensor was mounted at one end, and the gripper was attached to a slider. Impact tests were conducted by pulling the gripper at varying speeds via the tendon, resulting in different acceleration profiles. High-speed motion was recorded using a smartphone camera (OnePlus Ace Racing Edition; slow motion mode, 720p at 240 fps). The resulting footage was analyzed in MATLAB to extract the trajectory of the grasped object and correlate it with measured impact forces.

## 5. Discussion and Conclusion

In this work, we demonstrate that plastic deformation can be deliberately exploited as a functional mechanism for continuous passive stability in soft robotic grippers. By combining plastic deformation of an embedded metal layer with elastic recovery of the surrounding soft structure, the proposed continuously stable structure through plastic deformation (CSSPD) enables the gripper to retain a continuously adjustable configuration and grasping force after actuation is removed. This behavior differs from conventional bistable and multistable structures, which provide only discrete equilibrium states, and from actively stiffened systems that require continuous energy input to maintain their configurations. The resulting structure combines the compliance and adaptability of soft robotics with the passive shape retention and increased stiffness typically associated with rigid structures. The theoretical model further clarifies the relationship between plastic deformation, elastic recovery, and the retained bending configuration, and provides a basis for optimizing the kirigami geometry of the embedded metal layer.

The experimental results (Fig. S3 and Fig. S10), together with the fatigue-life analysis of the metal layer (Text S2), demonstrate the reliability of the structure under repeated actuation. The gripper maintains a passive holding force of up to 16 N without continuous energy input, while its enhanced stiffness enables stable operation under pulsed accelerations of up to 400 m/s $^2$when combined with pneumatic actuation. These results indicate that the CSSPD contributes to both static and dynamic grasp stability rather than merely providing passive shape retention. The proposed bioinspired paw pad further addresses the elastic recovery associated with plastic deformation by providing additional top-side support, thereby improving object stability during grasping. In combination with the Telescopic Soft Pressing Mechanism (TSPM), the paw pad also provides an integrated sensing and actuation function, enabling rapid object acquisition while maintaining mechanical support after grasp formation.

Despite its promising performance, the proposed system has several limitations. First, elastic recovery following plastic deformation is unavoidable and introduces a finite angular recovery after unloading. In the present design, this effect is partially compensated by the pre-stretched layer and the bioinspired paw pad. Further improvements in material selection and structural design could reduce the recovery angle and extend the fatigue life of the metal layer. In particular, metallic materials with more suitable combinations of yield strength, ductility, and fatigue resistance may provide improved long-term performance. Another potential approach is to integrate reconfigurable metallic materials, such as low-melting-point alloys, which could erase deformation history through thermally induced melting and resolidification, thereby enabling reversible structural reconfiguration. However, such approaches would introduce additional thermal energy requirements and response time and therefore involve a trade-off between reversibility, energy consumption, and actuation speed.

Second, the current prototype relies on pneumatic actuation to generate the initial deformation of the metal layer, requiring an external air pump and associated pneumatic equipment. This dependence limits direct integration into compact mobile robotic platforms. Alternative actuation approaches, including shape memory alloys, electroactive polymers, or magnetically responsive materials, could provide more compact implementations. However, these actuators would need to be co-designed with the embedded metal layer because their actuation force, displacement, response speed, and operating temperature directly influence the resulting plastic deformation and recovery behavior. Future optimization should therefore consider the actuator–structure interaction rather than treating the actuator and passive stability mechanism independently.

Further optimization of the kirigami geometry may also improve stability while reducing angular recovery. Although the current pattern is optimized primarily for finger bending, the underlying design principle is not restricted to a single deformation mode. By varying the kirigami topology, prestretch direction, and embedding configuration, the same principle could potentially be extended to expansion, contraction, twisting, or more complex multimodal deformations. Helically cut patterns, for example, could enable torsional or helical deformations, while multidirectional prestretched layers could facilitate more complex shape transformations. This geometric versatility provides opportunities for extending the CSSPD concept beyond the present gripper configuration.

The broader performance of the proposed gripper is summarized in Table S4, which compares the present system with existing approaches for stable grasping across multiple performance metrics. The comparison highlights the combination of continuous passive stability, retained grasping force, dynamic disturbance resistance, and low holding-energy requirements offered by the proposed architecture. In particular, the ability to maintain a grasp after actuation is removed provides an inherent degree of tolerance to actuator or pressure loss. In pneumatic systems, the gripper can retain the grasped object even after deflation, reducing the risk of accidental object release caused by sudden air leakage and thereby improving system safety and fault tolerance.

The demonstrated passive perching capability further suggests opportunities for mobile robotic applications. For aerial robots, the ability to maintain a stable grasp without continuous power could support high-speed interaction, tree perching, prolonged environmental monitoring, or docking for recharging. For underwater robots, passive retention could reduce energy consumption during long-duration grasping or docking under persistent current disturbances. In industrial environments, the combination of rapid sensing-based grasping and passive state retention could support high-throughput pick-and-place operations while reducing the energy required to maintain the grasp. More broadly, the results suggest that exploiting plastic deformation as a functional mechanical resource can provide a general design strategy for soft robotic systems that require a combination of compliance, continuous passive stability, dynamic robustness, and low energy consumption.

# Acknowledgment

**Funding:** This work was supported by AEA Ignite Fund, AE240300118.

# Supporting Materials

Figures S1 to S10

Tables S1 to S4

Texts S1to S3

Movies S1 to S10